# Advances and Challenges in Multimodal Remote Sensing Image Registration

Bai Zhu, Liang Zhou, Simiao Pu, Jianwei Fan, Yuanxin Ye, *Member, IEEE*

***Abstract* — Over the past few decades, with the rapid development of global aerospace and aerial remote sensing technology, the types of sensors have evolved from the traditional monomodal sensors (e.g., optical sensors) to the new generation of multimodal sensors [e.g., multispectral, hyperspectral, light detection and ranging (LiDAR) and synthetic aperture radar (SAR) sensors]. These advanced devices can dynamically provide various and abundant multimodal remote sensing images with different spatial, temporal, and spectral resolutions according to different application requirements. Since then, it is of great scientific significance to carry out the research of multimodal remote sensing image registration, which is a crucial step for integrating the complementary information among multimodal data and making comprehensive observations and analysis of the Earth's surface. In this work, we will present our own contributions to the field of multimodal image registration, summarize the advantages and limitations of existing multimodal image registration methods, and then discuss the remaining challenges and make a forward-looking prospect for the future development of the field.**

## I. INTRODUCTION

Resently, the multi-sensor integrated stereo observation technologies from spaceborne, airborne, and terrestrial platforms have been greatly developed with the launch of numerous remote sensing satellites and facilities, and it has completely entered the era of comprehensive intelligent photogrammetry [1]. Therefore, recent years have seen an explosion in the availability of multimodal image datasets in the remote sensing domain [2]. However, these multimodal remote sensing images (MRSIs) are usually acquired by different sensors with different imaging mechanisms in different periods for the same ground objects, such as optical-infrared, optical-multispectral, optical-synthetic aperture radar (SAR), and optical-light detection and ranging (LiDAR) (See Fig. 1), which generally gives rise to significant geometric distortion and radiometric differences among MRSI [3].

With the explosive growth of MRSIs, multimodal remote sensing image registration (MRSIR) is increasingly crucial because it is a prerequisite and pivotal step in plentiful co-processing and integration applications in the field of remote sensing, such as image fusion, change detection, and image mosaic (See Fig. 1). These co-processing and integration applications require the combined utilization of the above multimodal images to compensate for the deficiency of only using single-modal sensor image. Take the widely used panchromatic and multispectral images as an example. In the conflict between spectral resolution and spatial resolution, in order to obtain both high spatial resolution and spectral resolution images, image fusion is generally employed by fusing a high spatial resolution panchromatic and a low spatial resolution multispectral image one to obtain a multispectral image with the spatial resolution of the former while preserving the spectral information of the latter [4]. Similarly, there are complementary properties between optical and SAR [5], optical and LiDAR [6], and other multimodal images.

The pivotal step of MRSIR is to identify the correspondences with uniform distribution as much as possible between two or more MRSIs, the aim is to obtain the optimal geometric transformation model [7]. And registration accuracy has a crucial impact for the subsequent application even if the error of misregistration is only within several pixels or even a one-pixel range.

Nevertheless, the traditional registration methods are only suitable for homologous images with linear radiometric (intensity) differences, and also difficult to effectively extract common features between multimodal image pairs. These reasons lead to the difficulty of simultaneously detecting reliable correspondences and achieving satisfactory registration performance for MRSIs. Furthermore, although many general mainstreams commercial software (such as ENVI, ERDAS, PCI) have been developed internationally in which basic image registration function modules are provided, the traditional registration methods are still adopted to carry out the registration of multimodal image pairs. Recently, many advanced techniques such as invariant feature description and artificial intelligence have been employed for MRSIR in the remote sensing community because of the fast and robust performance requirement for MRSIR in those correlative co-processing and integration applications.

- This paper is supported by the National Natural Science Foundation of China (No. 42271446 and No.41971281), and in part by the Natural Science Foundation of Sichuan Province under Grant 2022NSFSC0537. (*Corresponding author: Yuanxin Ye*)
- Bai Zhu, Liang Zhou, Simiao Pu, and Yuanxin Ye are with the Faculty of Geosciences and Environmental Engineering, Southwest Jiaotong University, Chengdu 610031, China, and also with the State-Province Joint Engineering Laboratory of Spatial Information Technology for High-Speed Railway Safety, Southwest Jiaotong University, Chengdu 611756, China (e-mail: kevin_zhub@163.com; zhouliangmale@163.com; pusimiao_17@163.com; yeyuanxin@home.swjtu.edu.cn).
- Jianwei Fan is with the School of Computer and Information Technology, Xinyang Normal University, Xinyang 464000, China. (e-mail: fanjw@xynu.edu.cn)

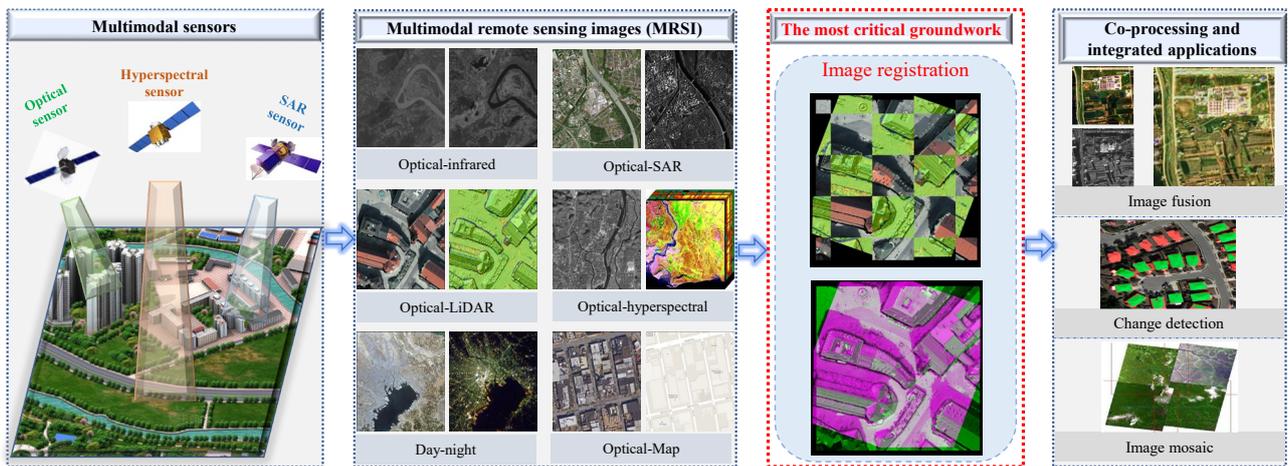

Fig. 1. Image registration is the most critical groundwork for co-processing and integration applications of MRSI.

In this paper, we will give an overview of the development of registration methods from unimodal to multimodal and our contribution to the field of MRSIR rather than delving into the details of specific algorithms or describing the results of comparative experiments in Section 2. Subsequently, we will then focus on the remaining challenges and future prospects in this field in Section 3.

## II. ADVANCES

As many researchers have suggested, the traditional MRSIR methodologies fall into two general categories with the taxonomy of area- and feature-based pipelines [8]. With the remarkable performance of deep learning on many complex tasks, it has also been used by many researchers in the field of image registration, and the learning-based pipeline has also developed into a set of methods that cannot be ignored [9, 10]. In the following, we will review the above three types of pipelines for performing multimodal registration in detail and comprehensively (See Fig. 2).

### A. Area-based Pipeline

The key component of the area-based framework is to select a suitable similarity metric that plays a significant role in the process of registration. It is necessary to use similarity metrics and optimization methods to accurately estimate geometric transformation parameters, so as to drive the optimization of the registration process. In general, the strategy of template matching is accompanied when area-based pipeline registers the sensed and reference image pairs under the guidance of a similarity metric, therefore, these methods are sometimes called template matching (See Fig. 3). The most important characteristic of remote sensing images compared with ordinary images is that the georeferencing of MRSIs can be obtained by advanced physical devices and navigation equipment, such as global position system (GPS) and inertial measurement unit (IMU). Therefore, area-based pipeline can be facilitated by utilizing the geo-referenced information of MRSIs because the offsets of the geo-referenced information for MRSIs typically range from several pixels to dozens of pixels in the image space [11].

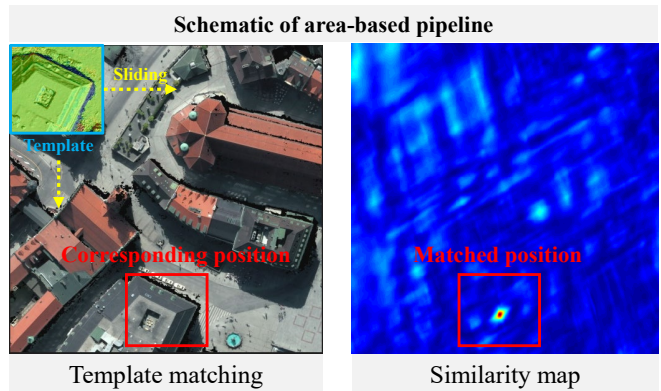

Fig. 3. Schematic of area-based Pipeline.

Area-based pipeline is generally composed of two independent domains: spatial domain and frequency domain, according to the different image representation domains. In the spatial domain, the sum of absolute differences (SAD), the sum of squared differences (SSD), the normalized cross-correlation (NCC), and the mutual information (MI) are the most representative similarity metrics. The most widely used similarity metric is phase correlation (PC) in the frequency domain, PC can address translation, scale, and rotation change between images to a certain extent by using the Fourier translation, rotation, and scaling properties [12].

However, traditional area-based registration methods usually evaluate the similarity of intensity information to find the optimal geometric transformation parameters that make the selected similarity metric reaches the maximum or minimum value to drive the iteration procedure. Therefore, SSD, NCC, and PC cannot provide robust registration results of MRSIs in most cases because these similarity metrics are sensitive to significant nonlinear radiometric (intensity) differences (NRD) that generally exist in multitudinous MRSIs [13]. And these above-mentioned metrics can only achieve robust registration performance in unimodal registration because the matching of correspondences is based on the computation of similar intensity values.

Among early area-based methods, information theory-based similarity metrics are the first one widely used for MRSIR and one of the most representative is MI, which is inspired by the field of medicine [14]. Subsequently, researchers have successively studied some novel solutions for MRSIR and made significant progress in both the accuracy and efficiency of these proposed algorithms. For nearly a decade, MRSIR has developed another representative pipeline in different categories, and we assign it to the generic name structure feature-based methods. We will respectively report a detailed description of these novel methods belonging to the above two categories in the following.

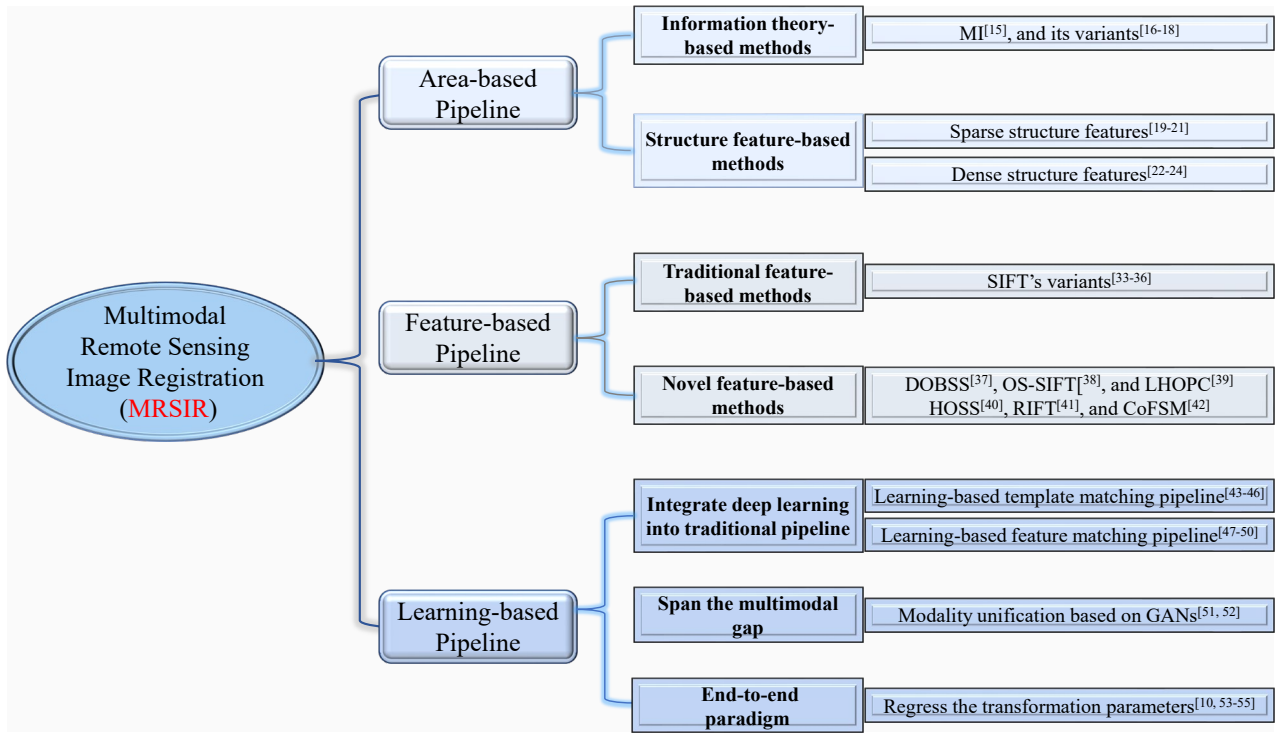

Fig. 2. Structure of this advances in multimodal remote sensing image registration (MRSIR).

Information theory-based methods are particularly suitable and extensively used for MRSIR because their statistical dependency can tolerate NRD to some extent. Moreover, they provide a more robust matching performance than other traditional metrics under the limitation that monomodal image registration has similar intensity values. Therefore, it has attracted great interest based on this basis to design advanced information-based metrics, such methods generally consist of the most common method to be MI [15], and its variants, such as the optimization of MI using a stochastic gradient [16], the Normalised Combined Mutual Information (NCMI) [17], and the symmetric form of MI [18].

Table I
Analysis of different area-based image registration methods

| Methods | | | Refer. | Year | Core idea | Advantages | Limitations |
|---|---|---|---|---|---|---|---|
| MI-based methods | | Improved MI | [15] | 2000 | Combine image pyramid approach and MI. | Address the NRD to some extent. | High computational complexity and lack of rotation invariance. |
| | | Optimized MI | [16] | 2003 | Optimize MI using a Stochastic Gradient. | Optimize the convergence speed of MI in terms of number of iterations. | High computational complexity and lack of scale and rotation invariance. |
| | | NCMI | [17] | 2014 | Adopt a non-parametric approach based on histogramming for joint probability density function estimation. | Improve the performance of NCMI as compared to conventional MI. | High computational complexity and lack of scale and rotation invariance. |
| | | Modified MI | [18] | 2016 | Use the symmetric form of Kullback-Leiber divergence as the similarity measure. | Improve the registration performance of multimodal image pairs without sufficient scene overlapping regions. | High computational complexity and lack of scale and rotation invariance. |
| Structure feature-based methods | Sparse | HOG | [19] | 2013 | Combine HOG and NCC. | Increases the probability and robustness of matching. | Lack of scale and rotation invariance. |
| | | DLSS | [20] | 2017 | Develop DLSC similarity metric by using NCC of the DLSS descriptors. | Address the NRD and show the superior matching performance. | High computational complexity and lack of scale and rotation invariance. |
| | | HOPC | [21] | 2017 | Define HOPCncc similarity metric by combining NCC with the HOPC descriptors | Address the NRD and achieve robust registration performance. | High computational complexity and lack of scale and rotation invariance. |
| | Dense | CFOG | [22] | 2019 | Present fast and robust template matching framework by integrating various local descriptors (HOG, LSS, and CFOG). | Resist the NRD and achieve fast and robust registration performance. | Lack of scale and rotation invariance. |
| | | OS-PC | [23] | 2020 | Propose subpixel registration method by combining robust feature representations of optical and SAR images and the 3D PC. | Show a superior performance in both accuracy and robustness | Lack of scale and rotation invariance. |
| | | SFOC | [24] | 2022 | Establish a fast similarity measure by employing FFT technique and the integral image. | Achieve superior matching performance in both registration accuracy and computational efficiency | Lack of scale and rotation invariance. |

Overall, those information theory-based methods are characterized by relying on the measure of statistical dependence and are suitable for MRSIR with different imaging mechanisms to some extent. However, the greatest limitation of information theory-based methods lies in their processes involved are very complicated and time-consuming, which restricts their wide application, especially for large-size multimodal image registration.

Structure feature-based methods are divided into sparse structure feature-based methods and dense structure feature-based methods according to whether they are pixel-by-pixel feature descriptions. Subsequently, the similarity of the generated features (sparse structure features or dense structure features) is evaluated by using classic metrics (e.g., SSD and NCC) to detect correspondences despite significant NRD, and noises between MRSIs. The former is commonly performed with geometric structure features extraction between MRSIs through a relatively sparse sampling grid, among which histogram of oriented gradients (HOG) [19], Dense Local Self-Similarity (DLSS) [20], and Histogram of Orientated Phase Congruency (HOPC) [21] are the most representative ones.

However, these above descriptors (i.e., HOG, HOPC, and DLSS) are generated by a relatively sparse sampling grid to capture structural or shape features. Therefore, they are all sparse representations, leading to some available and vital structural information being ignored, which further results in a depressed accuracy and robustness for the registration performance of these descriptors. While the dense structure feature-based methods are carried out by the pixel-wise feature representation that maps intensity information of images into a high-dimensional space by using descriptors, and representative methods include Channel Features of Orientated Gradients (CFOG) [22], feature representations of Optical and SAR images and the 3-D Phase Correlation (OS-PC) [23], and Steerable Filters of first- and second-Order Channels (SFOC) [24]. What's more, Ye et al. [24] constructed the SFOC descriptor that combines the first- and second-order gradients based on the steerable filters to capture more discriminative structure features of images. Furthermore, they used the FFT and integral images to accelerate the traditional NCC, and then developed a coarse-to-fine multimodal remote sensing image registration system on this basis, and provided a public code of their registration system.

For structure feature-based methods, it's worth noting that those sparse structure feature-based methods are only computed in the designed sparsely sampled grid, which means that they can only use the similarity metrics in the spatial domain to evaluate the similarity and identify correspondences, and the similarity metrics in the frequency domain are not applicable, such as PC. While the dense structure feature-based methods in another category can be used to match by taking advantage of the similarity metrics of both spatial and frequency domains. From the perspective of the robustness and efficiency of comprehensive registration, the dense structure feature-based methods have apparent effectiveness and advantages in the capacity of addressing significant NRD between MRSIs, which can meet many application requirements at present.

Table II
Analysis of different feature-based image registration methods

| Methods | | Refer. | Year | Core idea | Advantages | Limitations |
|---|---|---|---|---|---|---|
| Traditional feature-based methods | Improved SIFT | [33] | 2008 | Adopt various techniques including SIFT, Harris corner detector, the wavelet pyramid approach as well as cross matching and TIN construction strategy. | Improve the registration performance of multi-source image pairs with linear intensity differences. | It is difficult to perform robust registration of multimodal images with significant NRD. |
| | SR-SIFT | [34] | 2008 | Present scale restriction criteria for SIFT match. | Reduce the incorrect Matches for multi-spectral image registration. | It is difficult to perform robust registration of multimodal images with significant NRD. |
| | UR-SIFT | [35] | 2011 | Propose a selection strategy of SIFT features in the full distribution of location and scale where the feature qualities are quarantined based on the stability and distinctiveness constraints. | Applicable to various kinds of optical remote sensing images, even with those that are five times the difference in scale. | It is difficult to perform robust registration of multimodal images with significant NRD. |
| | AB-SIFT | [36] | 2015 | Apply an adaptive binning strategy to compute the local feature descriptor. | Resistant to a local viewpoint distortion and increase the robustness of matching. | It is difficult to perform robust registration of multimodal images with significant NRD. |
| Novel feature-based methods | DOBSS | [37] | 2015 | Present DOBSS descriptor consists of UR-SIFT feature extraction, correlation surface computation, orientation assignment using correlation surface histogram. | Against the illumination differences of the features in the multimodal images. | It is difficult to perform robust registration of multimodal images with significant NRD. |
| | OS-SIFT | [38] | 2018 | Propose an advanced SIFT-like algorithm (OS-SIFT) that consists of keypoint detection in two Harris scale spaces, orientation assignment and descriptor extraction, and keypoint matching. | Provide a robust registration result for optical-to-SAR images. | High computational complexity. |
| | LHOPC | [39] | 2018 | Derive the LHOPC feature descriptor by utilizing an extended phase congruency feature with an advanced descriptor configuration. | Show excellent performance especially for cases where there are complex radiometric differences. | High computational complexity. |
| | HOSS | [40] | 2019 | Calculate the self-similarity values in multiple directions using an oriented rectangular patch to increase the descriptor distinctiveness. | Increase the matching performance and address significant illumination variations. | Lack of scale invariance. |
| | RIFT | [41] | 2019 | Introduce MIM based on the log-Gabor convolution sequence for feature description. | Improve the robustness of descriptor to significant NRD. | Lack of scale invariance. |
| | CoFSM | [42] | 2022 | Present co-occurrence filter space and feature displacement optimization for multimodal matching. | Transform the NRD difference problem into the optimization of image feature similarity information. | High computational complexity. |

Among the area-based methods mentioned above, DLSS, [20], HOPC [21], CFOG [22], and SFOC [24] belong to our contributions in the field of MRSIR. Moreover, Table I provides a comprehensive analysis of the different area-based image registration methods. And publication year, core idea, advantages, and limitations of each method are listed in the table.

### B. Feature-based Pipeline

The area-based pipeline that has been explored above can only be applied to cases where there are only a few pixels or dozens of pixels offset between MRSIs with significant NRD, and they cannot effectively resist the influence of geometric distortions (e.g., scale and rotation changes). Although the georeferenced information is an inherent feature of remote sensing images and can be used to predict the approximate range for area-based pipeline, however, if the georeferenced information of images is unavailable, such as missing or the positioning error is too large (e.g., a few hundred pixels), the area-based pipeline cannot be carried out.

Different from area-based pipeline, feature-based pipeline usually does not rely on the georeferenced information of images as initial registration conditions. These methods usually detect significant salient and distinctive features (e.g., point features, line features, and region features) between images, then identify correspondences by describing the detected features. Therefore, feature detection and feature description are the most pivotal steps of this pipeline, which is also the core challenge of the multimodal feature matching method. Since the classical feature matching methods principally adopt the intensity or gradient information of images to perform feature detection and description, they can only be applied to the geometric distortions under the linear radiation difference.

In the feature matching method of the remote sensing field, the most widely used salient feature for detection is to extract distinctive interest points (IPs) with high repeatability, and the local region of IPs is described by constructing local invariant feature descriptors (See Fig. 4). Among classic methods, those IPs detection and description operators based on the intensity or gradient information of images are the most popular in computer vision. Moravec [25], Harris [26], Laplacian of Gaussian (LoG) [27], Differences of Gaussian (DoG) [28], and Features from Accelerated Segment Test (FAST) [29] are the most representative detection operator. The scale-invariant feature transform (SIFT) [28], the Gradient Location and Orientation Histogram (GLOH) [30], the Speed Up Robust Feature (SURF) [31], and DAISY [32] are widely used belong to the pipeline of local invariant features descriptors. Researchers in the field of remote sensing usually combine the characteristics of remote sensing images to modify or improve the above off-the-shelf operators for feature detection and description, such as the direct use of SIFT and Harris with other strategies [33], the Scale Restriction SIFT [34], the Uniform Robust SIFT [35], the Adaptive Binning SIFT [36].

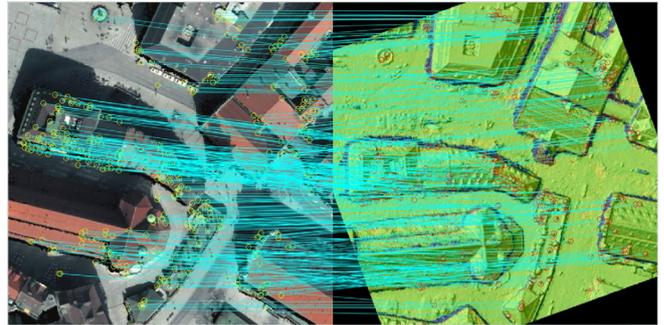
Fig. 4. Point features matching examples.

However, the above-mentioned feature detection and description operators are difficult to achieve excellent matching performance for MRSIs with both significant geometric distortions and NRD, because they are formed by utilizing the gradient or intensity information of images. In recent years, numerous efforts have been made in the field of MRSIR, and a whole range of novel feature-based methods have been proposed that can overcome the scale, rotation, radiance, and noise changes of MRSIs. In general, it is representative to design some salient features instead of gradients to obtain a stable feature description. These distinctive features include Distinctive Order Based Self Similarity (DOBSS) [37], an advanced SIFT-like algorithm (OS-SIFT) [38], and local histogram of orientated phase congruency (LHOPC) [39]. And they are more robust to radiometric differences than the descriptors based on gradient information.

Subsequently, a distinctive index map is proposed to obtain rotation invariance [40, 41]. Sedaghat and Mohammadi [40] explored a distinctive index map, called the rotation index of the maximal correlation (RIMC). And the RIMC is calculated by self-similarity values in multiple directions. Furthermore, The Radiation-Invariant Feature Transform (RIFT) was presented by Li et al. [41]. They constructed the maximum index map (MIM) by virtue of the log-Gabor convolution sequence and realized MIM's rotation invariance, which improves the stability of feature detection of MRSIs. However, both HOSS and RIFT only possess the rotation invariance and do not support the scale differences. Yao et al. [42] further constructed a novel co-occurrence scale space and log-polar descriptor that achieves both scale and rotation invariance for MRSIR.

Among the feature-based methods mentioned above, LHOPC [39] belongs to our contributions in the field of MRSIR. Moreover, Table II provides a comprehensive analysis of the different feature-based image registration methods. And publication year, core idea, advantages, and limitations of each method are listed in the table.

### C. Learning-based Pipeline

The rapid development of deep learning provides a new approach to address the problem of MRSIR, which has presented greater potential than traditional registration methods. Due to the powerful performance in nonlinear expression and deep feature generation, deep learning has attracted a lot of research attention in MRSIR. The methods of utilizing deep learning technology to perform image registration can be generally divided into following categories: (1) Integrate deep learning technology into the traditional pipeline to form a new registration process; (2) Span the multimodal gap by transforming one modality into another based on deep learning, thus simplifying MRSIR to monomodal remote sensing image registration; (3) Directly regress the transformation parameters between the multimodal images to realize an end-to-end MRSIR.

The most common strategy is to embed deep networks into the traditional pipeline to generate more robust features, more effective feature descriptors, or similarity metrics, thus ultimately improving the image registration performance. Most of the learning-based template matching methods are based on (pseudo-) Siamese networks [43, 44]. The basic idea of these methods is to obtain deep features by training a (pseudo-) Siamese network and then measure the similarity between the output feature maps to achieve image matching, which can be performed by neural network, dot product, convolution operation, or traditional similarity metrics. Considering the spatial information loss that may be caused by pooling operations, the fully convolutional architecture is used to generate feature maps of the same size as the input images [45, 46]. Learning-based template matching pipeline improves the registration performance, but they also cannot resist the geometric distortions (scale and rotation changes) between images.

Learning-based feature matching methods mainly use deep neural networks to perform feature detection and description and then measure the similarity between feature descriptors for MRSIR [47-49]. Zhao et al. [50] proposed a heterogeneous SuperPoint network, which operates on full-sized images and detects interest points accompanied with fixed length descriptors in a single forward pass. Learning-based feature matching pipeline can deal with MRSIs with scale and rotation differences, and the application of deep learning technology improves the repeatability of feature detection and the robustness of feature descriptors, so it has attracted more attention. However, the generalization of learning-based feature-matching methods needs to be further improved.

In addition, to address significant nonlinear radiometric differences between MRSIs, Generative Adversarial Networks (GANs) are introduced into MRSIR [51]. The core idea is to achieve modality unification based on GANs, so that the traditional registration methods can also achieve better registration results in the case of monomodal remote sensing image registration. These methods first perform image-to-image translation to span the multimodal gap by training GANs, making the MRSIs to be registered have similar radiometric or feature information. Then, area-based or feature-based methods are carried out, which effectively enhance the performance of image registration. Merkle et al. [52] used conditional GANs to generate pseudo-SAR image patches from optical images and then effectively improve the matching accuracy and precision of traditional approaches including NCC and SIFT based on the artificial patches.

Table III
Analysis of different learning-based image registration methods

| Methods | | Refer. | Year | Core idea | Advantages | Limitations |
|---|---|---|---|---|---|---|
| Integrate deep learning technology into the traditional pipeline | × | [43] | 2017 | Employ dilated convolution to expand the receptive field and adopt the dot product layer to measure the similarity between features. | Larger receptive field. | Depend on the salient features in the image scene. |
| | × | [44] | 2018 | Adopt a fully connected layer to fuse the features and apply a loss function based on binary cross entropy. | Solve the matching task in very high-resolution optical and SAR images. | The fully connected layer causes the loss of spatial information. |
| | SFcNet | [45] | 2019 | Propose a Siamese fully convolutional network and train by maximizing the feature distance between positive and hard negative samples. | More effective training strategies. | There is still spatial information loss. |
| | MCGF | [46] | 2021 | Employ the deep learning technique to refine structure features for improving image matching. | Capture finer common features between SAR and optical images. | Lack of scale and rotation invariance. |
| | DescNet | [47] | 2019 | Train a deep neural network to generate a robust feature descriptor for feature matching and apply the hardest sample mining strategy. | Acquire more matched points. | Lack of radiation invariance. |
| | LSV-ANet | [48] | 2021 | Transform outlier detection into a dynamic visual similarity evaluation. | Allow structure manipulation and scale selection of descriptors within the network. | Lack of radiation invariance. |
| | M2DT-Net | [49] | 2022 | Combine learning features and the Delaunay triangulation constraint. | Achieve a balance of robustness and time consumption. | The robustness of the model needs to be further improved. |
| | × | [50] | 2022 | Operate on full-sized images and produces multi-modal interest points accompanied with fixed length descriptors in a single forward pass. | Improve the performance under different scales and rotation changes | The complexity of the method needs to be further reduced. |
| Simplify MRSIR to monomodal remote sensing image registration | GMN | [51] | 2018 | Propose a generative matching network to generate the coupled optical and SAR images. | Improve the quantity and diversity of the training data. | The quality of the generated images needs further improvement. |
| | × | [52] | 2018 | Train conditional generative adversarial networks (cGANs) to generate SAR-like image patches from optical images. | Provide a new concept to handle the problem of MRSIR based on cGAN | The quality of the generated images needs further improvement. |
| Directly regress the transformation parameters | × | [53] | 2020 | Two convolutional neural networks are cascaded to realize the unsupervised registration framework. | Achieve good registration performance and processing efficiency. | Lack of radiation invariance. |
| | × | [54] | 2021 | Utilize different time steps to refine and regress displacements with an iterative way. | The method can be integrated into any kind of fully convolutional architecture. | Lack of radiation invariance. |
| | × | [55] | 2018 | Design fully convolutional neural networks to learn scale-specific features. | Get rid of gradient descent schemes by predicting directly the deformation. | Lack of radiation invariance. |
| | MU-Net | [10] | 2022 | Stack several deep neural network models on multiple scales to generate a coarse-to-fine registration pipeline. | Achieve good registration performance between image pairs with geometric and radiometric distortions. | Depend on the overlap area between image pairs. |

Most of the above methods need to be equipped with mismatch removal and parameter estimation algorithm to optimize matching results and obtain the final transformation parameters between images, thus are not end-to-end image registration frameworks. At present, more and more researches focus on directly estimating geometrical transformation parameters or deformative field, but are mainly applied in the vision and medical fields. Some works introduce this idea into the remote sensing community [53, 54]. It can be further divided into supervised and unsupervised methods depending on whether ground truth is used. Focused on the nonrigid registration problem, Zampieri et al. [55] built a fully convolutional network to obtain scale-specific features, which are easy to train and can realize nonrigid MRSIR in linear time. Ye et al. [10] proposed an unsupervised network, named MU-Net, to directly learn the transformation parameters with a multiscale manner. MU-Net forms a coarse-to-fine registration framework by stacking several deep networks on multiple scales and designs the loss function based on structural similarity to make it suitable for multimodal images, which can resist both geometric and radiometric distortions.

Among the learning-based methods mentioned above, multiscale convolutional gradient features (MCGFs) [45] and MU-Net [10] belong to our contributions in the field of MRSIR. Moreover, Table III provides a comprehensive analysis of the different learning-based image registration methods. And publication year, core idea, advantages, and limitations of each method are listed in the table.

## III. CHALLENGES AND PERSPECTIVES

Image matching or registration for MRSIs has played a crucial role in a variety of key applications (including image fusion, change detection, and image mosaic) in the field of remote sensing. Over the past decades, researchers have explored many new methods in this field, and the accuracy and efficiency of MRSIR have been constantly improved. In this paper, the related research in this field is comprehensively summarized with the aim of providing an informative reference for relevant readers. This summary separately reviews the classical and up-to-date pipelines of three different categories consisting of the area-, feature-, and learning-based pipelines.

Despite different matching pipelines having made great progress in both methodology and performance with the efforts of researchers, there are certain limitations in the implementation of different pipelines, and they are not suitable for MRSIR in various complex situations. Therefore, we will discuss each of these questions for different pipelines in series.

The limitations of area-based pipeline are mainly manifested in two aspects. First of all, all these methods depend on some prior conditions to predict the approximate matching range in order to further implement the subsequent template matching process. These prior conditions can be either the geo-referenced information of MRSIs or a rough mapping model (e.g., affine and projective transformation model) that can be obtained by manually selecting a few checkpoints with uniform distribution. Another important limitation is that area-based pipeline has requirements for the overlapping area between the sensed and reference image. If the overlapping area of the sensed and reference image is too small, the robustness

Table IV
Detailed information of current public datasets

| Datasets | Image type | Year | Resolution (m) | Image Size (Pixels) | Number of image pairs | Download link |
|---|---|---|---|---|---|---|
| RGB-NIR Scene dataset [56] | visible-infrared | 2011 | unknown | multiple | 477 | https://ivrlwww.epfl.ch/supplementary_material/cvpr11/index.html |
| TNO dataset [57] | multiple | 2017 | multiple | multiple | × | https://figshare.com/articles/dataset/TNO_Image_Fusion_Dataset/1008029 |
| SARptical [58] | optical-SAR | 2018 | 1 | 112*112 | 10,108 | https://syncandshare.lrz.de/getlink/fiGixjRV9idETzPgG689dGB/SARptical_data.zip |
| SEN1-2 [59] | optical-SAR | 2018 | 10 | 256*256 | 282,384 | https://mediatum.ub.tum.de/1436631 |
| SEN12MS [60] | optical-SAR | 2019 | 10 | 256*256 | 270,993 | https://mediatum.ub.tum.de/1474000 |
| RoadScene dataset [61] | visible-infrared | 2020 | unknown | multiple | 221 | https://github.com/jiayi-ma/RoadScene |
| OS-dataset [62] | optical-SAR | 2020 | 1 | 256*256 512*512 | 10,692 2,673 | https://pan.baidu.com/share/init?surl=4bqaJhMSZEy7EXcXVAc77w (code: vriw) |
| QXS-SAROPT dataset [63] | optical-SAR | 2021 | 1 | 256*256 | 20,000 | https://github.com/yaoxu008/QXS-SAROPT |
| DIML/CVL RGB-D dataset [64] | optical-depth | 2021 | multiple | multiple | 476,000 | https://dimlrgbd.github.io/ |
| LLVIP dataset [65] | visible-infrared | 2021 | unknown | 1920 * 1080 1280 * 720 | 16,836 | https://bupt-ai-cz.github.io/LLVIP/ |
| Multimodal_Image_Matching_Datasets [8] | multiple | 2021 | multiple | multiple | 164 | https://github.com/StaRainJ/Multi-modality-image-matching-database-metrics-methods/blob/master/Multimodal_Image_Matching_Datasets/ |
| MRSI Datasets [42] | multiple | 2022 | multiple | multiple | 46 | https://skyearth.org/publication/project/CoFSM/ |

of the matching will be greatly affected and even lead to the failure of the matching.

Feature-based pipeline universally constructs high-dimensional informative feature vectors for each of IPs by exploiting the local spatial relationships among their neighboring patches, and these high-dimensional informative feature vectors are very crucial for the matching performance of those proposed descriptors. As a result, the feature matching methods of this pipeline usually face a heavy computational burden, and are prone to produce inevitable heavy outliers (mismatches) in putative matches, especially appear in the case of multimodal registration with simultaneous scale, rotation, and radiation differences. In general, the robustness of the feature-based pipeline is not as stable as that of the area-based pipelines.

Learning-based pipeline is completely driven by training data, which can effectively enhance the performance of image registration. However, learning-based pipeline proposed for MRSIR is not as rich as those in the vision and medical community. The main reason is that multimodal remote sensing images are more complex than natural and medical images because of diverse scenes, wide imaging range, and hybrid noise, which results in it quite difficult to construct a sufficient and representative dataset for model training and testing. Moreover, due to the higher acquisition cost or the avoidless secrecy requirement, only a small number of remote sensing image registration datasets are publicly available (Table IV provides the detailed information of some current public datasets, including download links), which greatly limits the development of learning-based pipeline for MRSIR.

In summary, there are still many remaining unresolved issues in MRSIR, A general method for the robust and fast MRSIR remains an open problem for both handcrafted and learning pipelines, and future development will mainly focus on the following aspects.

The research of area-based pipeline mainly focuses on solving the significant NRD between MRSIs, and the research on the template region between the sensed and the reference image with both severe NRD and complex deformation (such as affine deformation) is relatively scarce. Future research could focus on the exploration of deformable similarity metric, which combines local appearance with structure features to achieve reliable matching of multimodal images with both severe NRD and complex deformation.

The research of feature-based pipeline is more inclined to design stable operators for feature detection and feature description to seek dependable correspondences of MRSIs with scale, rotation, and radiation differences. Future research can be explored from the following two aspects. On the one hand, researchers can focus on exploring acceleration strategies (such as the FFT technique) for feature detection and description to improve the computational efficiency of the pipeline. On the other hand, future work predominantly will explore outlier removal ways (such as the stable neighborhood topologies and piecewise linear transformation) to robustly eliminate false correspondences from putative correspondences identified on the basis of descriptor similarity.

Suffer from the lack of datasets, it is difficult to obtain the network model with good generalization for learning-based pipeline. Future works should pay more attention to providing more publicly available datasets. It can be achieved by collecting more multimodal images based on open-source satellite platforms to generate new datasets. In addition, unsupervised learning is a potential way to achieve better registration results under limited datasets, which is worth more exploration in the field of MRSIR. More advanced concepts (e.g., contrastive learning) and more efficient feature extraction models can be better combined for MRSIR.